\documentclass{article}

\usepackage[preprint]{neurips_2023}

\usepackage[utf8]{inputenc} % allow utf-8 input
\usepackage[T1]{fontenc}    % use 8-bit T1 fonts
\usepackage{hyperref}       % hyperlinks
\usepackage{url}            % simple URL typesetting
\usepackage{booktabs}       % professional-quality tables
\usepackage{amsfonts}       % blackboard math symbols
\usepackage{nicefrac}       % compact symbols for 1/2, etc.
\usepackage{microtype}      % microtypography
\usepackage{xcolor}         % colors
\usepackage{amsmath}
\usepackage{graphicx}

\title{Linear Latent World Models in Simple Transformers: A Case Study on Othello-GPT}

\author{
Dean S. Hazineh$^{*}$  \\
Department of Applied Physics \\
Harvard University \\
\texttt{dhazineh@g.harvard.edu} \\
\And
Zechen Zhang$^{*}$ \\
Department of Physics \\
Harvard University \\
\texttt{zechen\_zhang@g.harvard.edu} \\
\AND
Jeffrey Chiu \\
Department of Computer Science\\
Harvard University \\
\texttt{jeffrey\_chiu@college.harvard.edu}
}

\begin{document}

\maketitle

\begin{abstract}
Foundation models exhibit significant capabilities in decision-making and logical deductions. Nonetheless, a continuing discourse persists regarding their genuine understanding of the world as opposed to mere stochastic mimicry. This paper meticulously examines a simple transformer trained for Othello, extending prior research to enhance comprehension of the emergent world model of Othello-GPT. The investigation reveals that Othello-GPT encapsulates a linear representation of opposing pieces, a factor that causally steers its decision-making process. This paper further elucidates the interplay between the linear world representation and causal decision-making, and their dependence on layer depth and model complexity. We have made the \href{https://github.com/DeanHazineh/Emergent-World-Representations-Othello}{code} public.  \footnotetext{* Equal contribution}
\end{abstract}

\section{Introduction}
Auto-regressive language models, esteemed for their predictive capabilities, have showcased unprecedented potential in logical reasoning and complex task resolution (\cite{srivastava2022beyond}\cite{creswell2022selection}). This leads to a discourse (\cite{bender2021dangers}) regarding whether their performance is merely a reflection of memorization of “surface statistics,” namely the correlations between tokens unrelated to the original sequence's causal process. Contrarily, recent research endeavors examine the inherent development of interpretable world models within such language models. World models, as referred throughout, signify the encoding of semantic information about the true causal process within the network’s activations. Instances of world representations include the location and type of game pieces for a language model trained in chess (\cite{toshniwali2021}) and object part classification in a visual transformer (\cite{amir2021}). Exploring the unsupervised development of internal world representations in these models is paramount for comprehending the emergence of intelligence in foundation models.

In this research, the emergent world principle is scrutinized in a series of simple transformer models, trained to play Othello and termed as Othello-GPT (Appendix \ref{appendix:rules}). These trained models (Figure \ref{fig:GPT_and_Probe_Schematic}a) exhibit proficiency in legal move execution. Utilizing linear probes to decode neuron activations across transformer layers, coupled with causal interventions, this paper underscores the enhanced world model of Othello-GPT and its causal utility for decision-making.

\paragraph{Contributions:}
In this ongoing work, 
\begin{enumerate}
    \item We show that even very small models down to 1 layer and 1 attention head can play Othello, and they possess \textbf{linearly} encoded information about the board state. We find that world representations may be linearly encoded in the model's activations to an extent that generally increases with layer depth for deeper models. 
    \item We design a simple novel causal intervention technique that directly intervenes at each layer with the linear representation and demonstrate that at certain layers the linear world model is \textbf{causal} in influencing the model's next move prediction. In particular, it appears that semantic understanding in the model is developed and utilized about halfway through the model.
    
\end{enumerate}

\section{Related Work}
\paragraph{Othello-GPT}
Li et al. (\cite{li2023emergent}) inaugurated the research on an 8-layer transformer model for Othello, demonstrating the non-linear encoding of board state information within each layer's activations and introducing a sophisticated causal intervention scheme. A subsequent blog post (\cite{nanda_othello_2023}) highlighted the potential for linear encoding of the board state, albeit without an interpretable representation.

\paragraph{Probing Classifiers}
The employment of classifiers for probing neural networks has gained prominence as an instrumental tool for understanding internal representations within neural networks. This methodology has facilitated the understanding of sentence embeddings in transformers (\cite{clark-etal-2019-bert}), hierarchical structures in recurrent networks (\cite{hupkes2018visualisation}), and visual concepts within vision models (\cite{alain2018understanding}). Despite these advancements, debates continue (\cite{belinkov2021probing}) regarding the actual causal utility of the encoded information by the network. This research contributes a meticulously designed causal intervention study to ascertain the network's utilization of the encoded representation.

\paragraph{Mechanistic Interpretability}
Complementary research (\cite{olah2017feature} \cite{olah2020zoom}\cite{olsson2022incontext}\cite{nanda_othello_2023}) endeavors to decipher the internal mechanism of networks by identifying "circuits," network segments responsible for computing certain interpretable features. This paper lays a robust foundation for further mechanistic interpretability work on Othello-GPT.

%%%%%%%%%%%%%%%%%%%%%%%%%%%
%%% Section 1
%%%%%%%%%%%%%%%%%%%%%%%%%%%
\section{Linear Representation of the Board State}
\label{sec:world_representation}

\begin{figure}[t!]
    \centering
    \includegraphics[width=1.0\columnwidth]{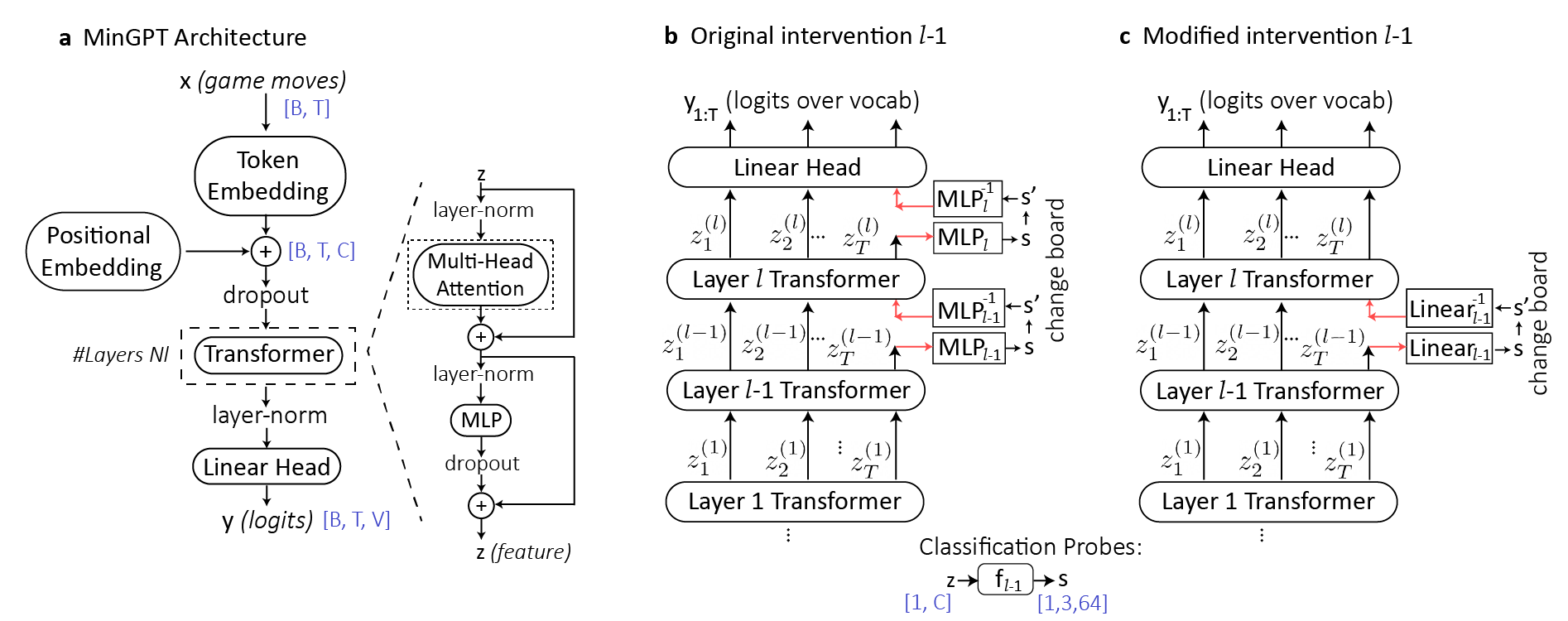}
    \includegraphics[width=1.0\columnwidth]{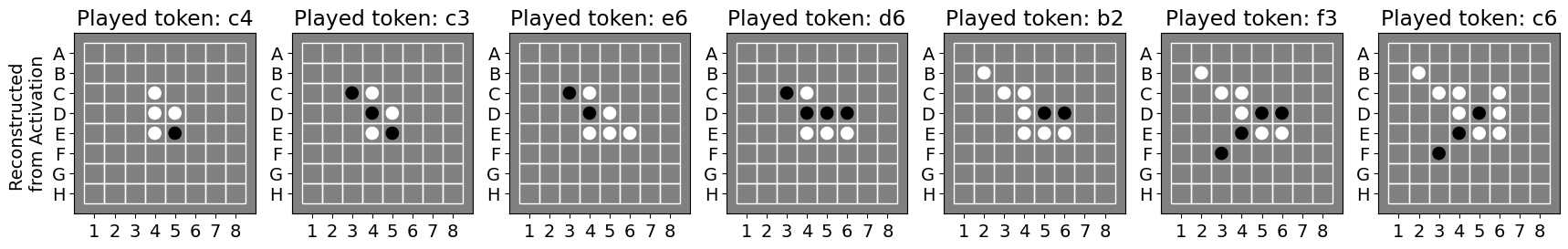}
    \caption{Overview of the principles in section \ref{sec:world_representation}. (a) The neural architecture utilized in this paper, where the number of layers refers to the number of transformer blocks. (b) The original intervention scheme of \cite{li2023emergent} is replaced by an alternate version shown in (c), whereby intervention is applied to a single layer--see text for details. (Bottom Panel) Example of an extracted world representation where the game board state is obtained from the activation vectors.}
    \label{fig:GPT_and_Probe_Schematic}
\end{figure}

\subsection{Linear Probe Uncovers a Natural Board State Representation}
\label{ssec:linear-world-representation}

In this section, we first report our findings that there is an intuitive world representation that is linearly encoded within the activation vectors of the trained sequential model. 

The original work of \cite{li2023emergent} postulated a world representation corresponding to the placement of white or black discs on a game board. While this is how most humans would visualize the board when playing the game of Othello, it is important to note that the sequential model differs slightly from human play in that the model alternates between acting as the white vs. black player depending on the length of the input sequence. As the rules dictating a legal move require only an understanding of which piece is yours and the opponents', a more natural choice for a world representation in this case is instead the classification of a tile as containing "my piece" vs. "your piece" (or empty). We term the "Black, White, Empty" representation as the old representation, and "Yours, Mine, Empty" as the new representation.

To quantify this, we first train a multi-headed attention model (diagrammed in Figure \ref{fig:GPT_and_Probe_Schematic}a) with 8 transformer blocks, referred to as layers, and 8 attention heads on the synthetic Othello dataset that consists of random legal moves. Once trained to predict legal next moves, the model's weights are frozen. We then train a separate linear transformation $L$  (referred to throughout as a linear probe) to map the activation vector of a particular layer to the world representation, i.e., the mapping $f: \mathbb{R}^{C=512} \rightarrow \mathbb{R}^{64x3}$ where $C$ is the token embedding dimension and the output corresponds to the three-state classification logits for each of the 64 tiles. The accuracy for the linear probe at each layer for the old vs. new world representations is shown in Table. \ref{table:probes}. We observe that the new representation is linearly encoded in the activation vectors, particularly within the deeper layers, and can be recovered from the model's activations by a simple linear transformation with near unity accuracy on a withheld set. Surprisingly, the linear probe is successful even for a 1-layer Othello-GPT, as will be discussed in Appendix \ref{appdix: shallow}.

\begin{table}[h!]
    \centering
        \caption{Classification Accuracy for Linear Probes Mapping $z\rightarrow s$}
        \begin{tabular}{  c | c c c c c c c c  }
        \hline
        Layer: & 1 & 2 & 3 & 4 & 5 & 6 & 7 & 8\\
        \hline 
         Old (Black,White)  & 75.7$\%$ & 75.8$\%$ & 75.7$\%$ & 75.7$\%$ & 75.6$\%$ & 75.4$\%$ & 74.9$\%$ & 74.9$\%$  \\
       New (Mine, Yours)& \textbf{90.8}$\%$ & \textbf{94.8}$\%$ & \textbf{97.1}$\%$ & \textbf{98.3}$\%$ & \textbf{99.1}$\%$ & \textbf{99.5}$\%$ & \textbf{99.5}$\%$ & \textbf{99.5}$\%$ \\
        % 9 & 10 & 11 & 12 \\
        % 13 & 14 & 15 & 16 \\
        % 17 & 18 & 19 & 20 \\ [1ex]
        \hline
        \end{tabular}
    \label{table:probes}
\end{table}

\subsection{"Yours" and "Mine" Attention Heads}
\label{section:att}
% Note Dean: I think you can merge this into the previous section to save space
\begin{figure}[t!]
    \centering
    \includegraphics[width=0.8\columnwidth]{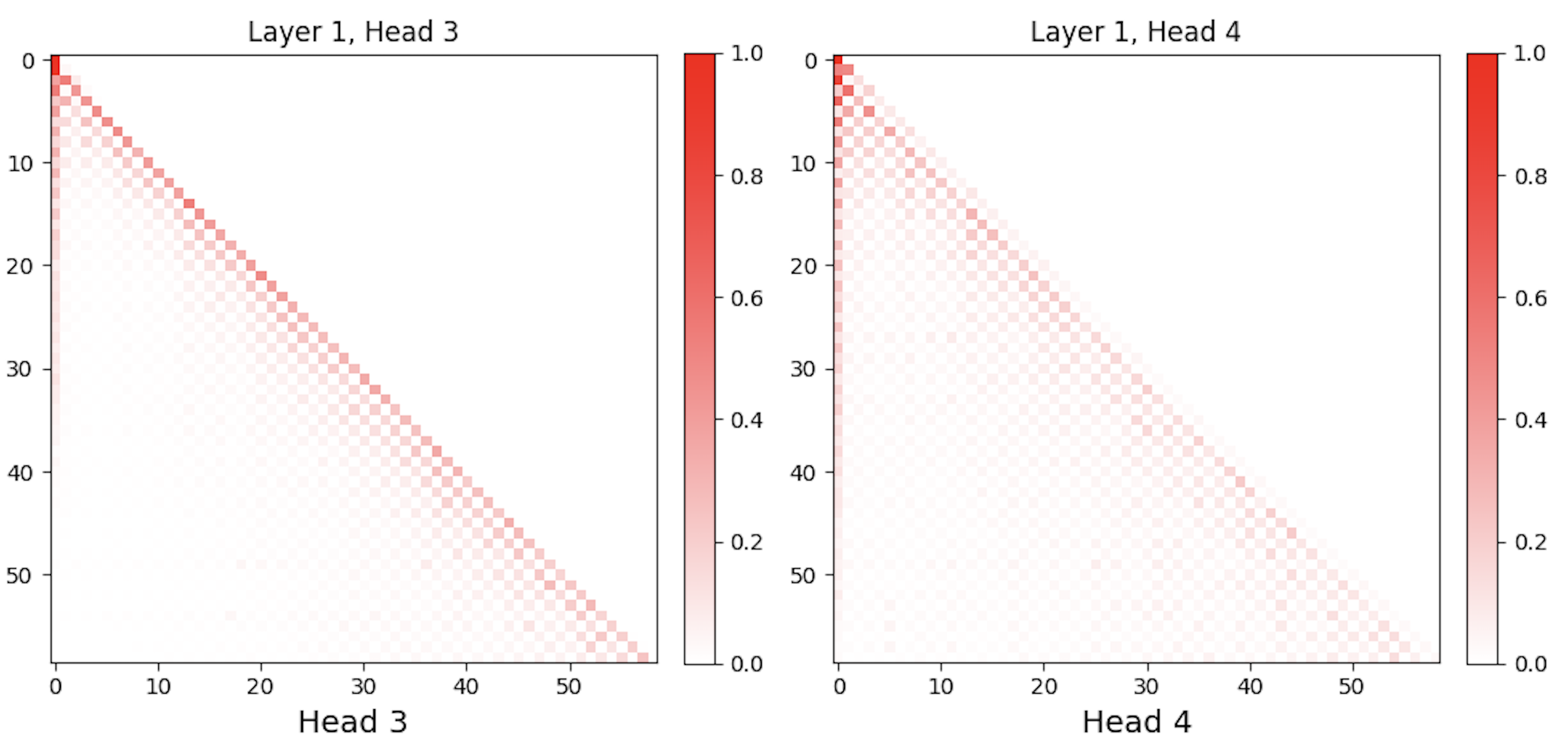}
    
    \caption{Example attention heads of the 8L8H model. Note that both heads show alternating patterns, which we interpret as processing information from pieces first placed by the same player. For example, Head 3 is keeping track of all my historical moves and Head 4 is keeping track of all the opponent's historical moves. }
    \label{fig:1L8H_attn}
\end{figure}

In addition to the linearly encoded board states that represent "yours" and "mine" pieces, we also find regularities in the attention heads that support such representation. For example,  Figure \ref{fig:1L8H_attn} shows that there exist attention heads that only keep track of "your" first-placed pieces and "my" first-placed pieces, resulting in alternating patterns. In fact, the "yours" and "mine" attention heads occupy a large percentage of the attention heads in the 8L8H Othello-GPT, and these attention heads emerge even in a simple 1L transformer (Appendix \ref{appendix:8L_attn}). 

\section{Causal Interventions}
To see if the linear world model is used by the Othello-GPT in making next-move predictions, we intervene at each layer of the model at certain game length, to trick the model into believing that it has another board state, as shown in Figure \ref{fig:GPT_and_Probe_Schematic}c. In general, we take the activations $z$ that corresponds to the board state $s$ per the linear probe, then change it to some other board state $s'$ and map it back to an activation $z'$ that is associated with it, ie. 
\[
L(z') =s',
\]
where $L$ is the linear transformation induced by the linear probe and we use gradient descent on a linear MLP to find an inverse map $L^{-1}$.
\subsection{Latent Saliency Map}
\begin{figure}[t!]
    \centering
    \includegraphics[width=1.0\columnwidth]{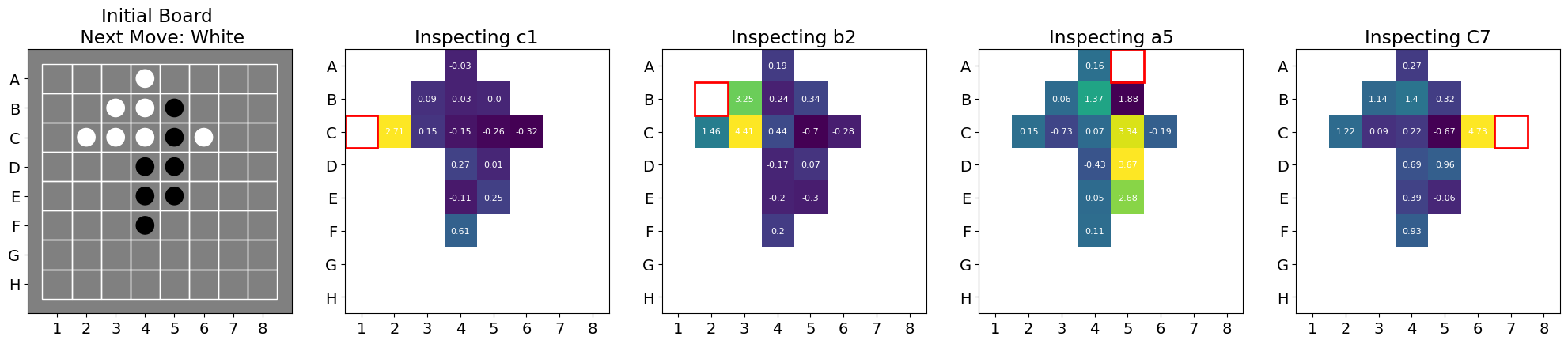}
    \caption{Latent saliency map for a particular game, interventions at layer 6. $m^*$ is highlighted with a red box which indicates the move that we calculate logits for and the numbers indicate the logit changes given interventions at that tile. If a possible next move is currently illegal, we observe that interventions that would make the move legal produce a positive change (bright yellow); alternatively, interventions that make a currently legal move now illegal have a negative change (dark blue).}
    \label{fig:Latent_saliency}
\end{figure}

First, we demonstrate that the interventions from changing the activation $z$ to $z'$ is indeed causal in certain layers by plotting the latent saliency map of interventions. The map is obtained by flipping each piece at a certain game length and recording the change of logits of a particular move $m^*$. If the intervention is indeed causal, we would expect that some particular interventions would encourage or discourage the move $m^*$, and this is indeed what we observe in Figure \ref{fig:Latent_saliency}. It is worth reporting that the interventions are successful only at certain layers, which we will discuss more in the next section.

\subsection{When and Where Causal Intervention is Successful}
\label{ssec:causal}
\begin{figure}[h]
    \centering
    \includegraphics[width=1.0\textwidth]{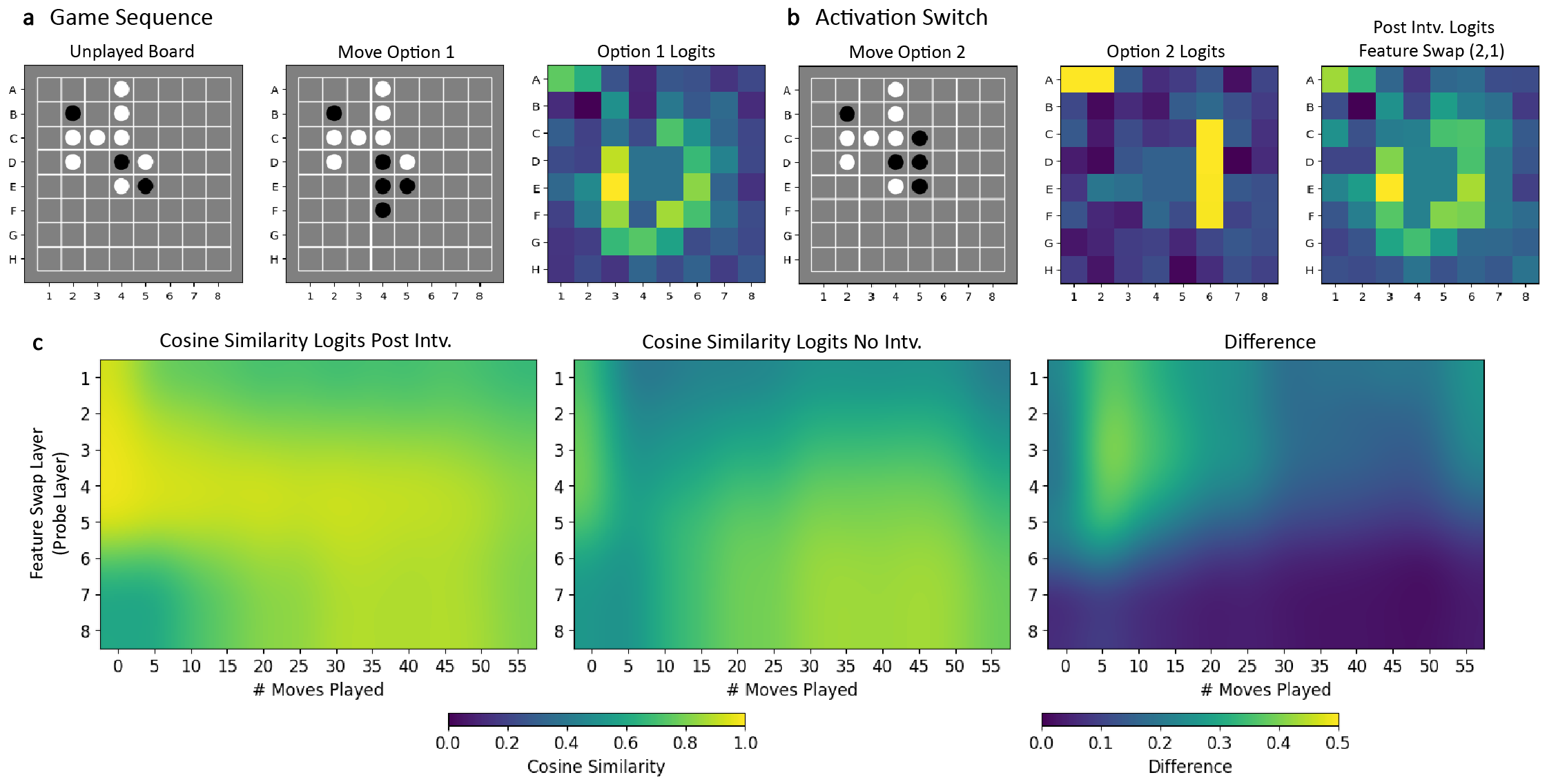}
    \caption{Comparison of logit distributions with causal interventions. (a) Logit distribution for move option 1 at a particular game length (unplayed board). (b) Logit distribution of move option 2 pre- and post-interventions. The logit distribution post intervention resembles that of move option 1. (c) Cosine similarities between the two options pre and post interventions, averaged over 50 sample games. The difference between the the two cosine similarities shows when and where intervention is most influential.}
    \label{fig:cos_sim}
\end{figure}

The latent saliency map is limited as it is only a qualitative demonstration and is also restricted to flipping-piece interventions. To see when (game length) and where (layer) the linear board state is used by the model causally, we look at the logit distribution of a model tricked into believing it has another board state. Specifically, at game length $t$ (termed unplayed board), we introduce two move options $s_1,s_2$ and intervene with the second move option at each layer to change activations $z_2$ to $z_1$ that corresponds to the board state $s_1$. If the intervention is indeed causal, then we would see that the post-intervention logit distribution of move option 2 is very similar to that of move option 1, as shown in Figure \ref{fig:cos_sim}a,b.

We plot the cosine similarity between the logit distribution of move option 1 and that of move option 2 post-intervention in \ref{fig:cos_sim}c, with cosine similarity of the two move options without interventions as a reference. We find that the intervention is mostly successful in middle layers and in early games. In particular, it seems that the intervention at the last two layers is not causally affecting the next-move decision at all, which is surprising since the linear probe has near peak accuracy at last few layers according to Table \ref{table:probes}. In addition, in appendix \ref{appdix: shallow} we show that shallow networks are worse at using the board state representation causally.

Taking account of the attention head distribution (Appendix \ref{appendix:8L_attn}), we hypothesize that the linear world model is fully developed in the middle layers and the residual stream is mostly using the information in the middle layers to make next-move decisions. In late games, however, Othello-GPT might only rely on "surface statistics" as the game tree is vastly shrinking. 

\section{Conclusion}
In this paper, we showed that the Othello-GPT neurons do encode information of the board state linearly, even for a shallow 1-layer network. Furthermore, the board state representation is used to make predictions with deep networks but not with shallow networks. Lastly, we demonstrated that the linear board state information is already finalized in the middle layers of the Othello-GPT and is used casually for decision-making, indicating that the transformer recalls information in the middle layers to make predictions. 

In general, our work provides a firm playground for future explorations on understanding when, where, and how foundation models utilize the internal world model to make predictions. Future work involves using mechanistic interpretability tools to understand the circuits within the Othello-GPT and breaking down its decision-making process.

\section{Acknowledgement}
The authors would like to thank Neel Nanda and Kenneth Li for open-sourcing the Othello-GPT models. ZZ is supported by the Center for Brain, Minds, and Machines at MIT.
%%%%%%%%%%%%%%%%%%%%%%%%%%%
%%% Appendix
%%%%%%%%%%%%%%%%%%%%%%%%%%%
\appendix
\section{The Rules of Othello and Board Represntations}
\label{appendix:rules}
Here, we provide a brief summary of the rules of the game. Othello, also known as Reversi, is a classic strategy board game for two players, typically played on an 8x8 grid with two-sided pieces called "discs" that are black on one side and white on the other. The objective is to have more discs of your color on the board than your opponent at the end of the game. All games start with an initial setup of four discs placed at the center of the board, with two white discs forming a diagonal and two black discs forming the opposite diagonal. One player will play black discs while the other will play white. Players then take turns placing one disc of their color on an empty square on the board, with the white player always going first.
\begin{itemize}
\item Flipping discs: A move is valid only if it "sandwiches" (or "flanks") one or more of your opponent's discs between the disc you are placing and another of your discs already on the board. This must occur in a straight line (horizontal, vertical, or diagonal). After placing the disc, flip all the sandwiched opponent's discs to your color. If a player cannot make a valid move, they must pass their turn. 
\item Win Condition: The game ends if neither player can make a valid move or if the board is full. The player with the most discs of their color on the board wins the game. If both players have the same number of discs, the game is a draw.

% The old (Black, White) representation and the new (Mine, Yours) representation are presented in Figure.
\end{itemize}

\section{Emergence of Linear Representations in Smaller Othello-GPT Models}
\label{appdix: shallow}

% \begin{figure}[t!]
%     \centering
%     \includegraphics[width=1.0\columnwidth]{Figures/GPT_Model.png}\\
%     \caption{[94.9$\%$, 98.6$\%$, 99.7$\%$, 99.6$\%$, 99.9$\%$]}
%     \label{fig:trained_GPT_Models}
% \end{figure}
\begin{figure}[t!]
    \centering
    \includegraphics[width=1.0\columnwidth]{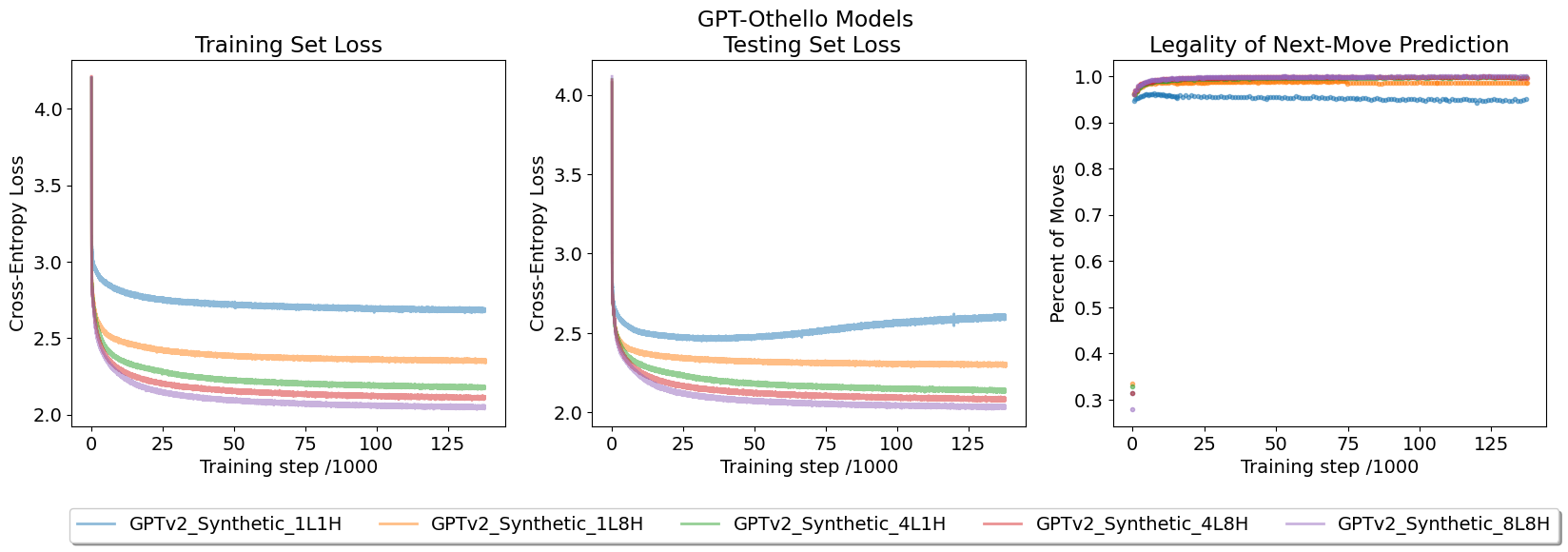}\\
    \includegraphics[width=1.0\columnwidth]{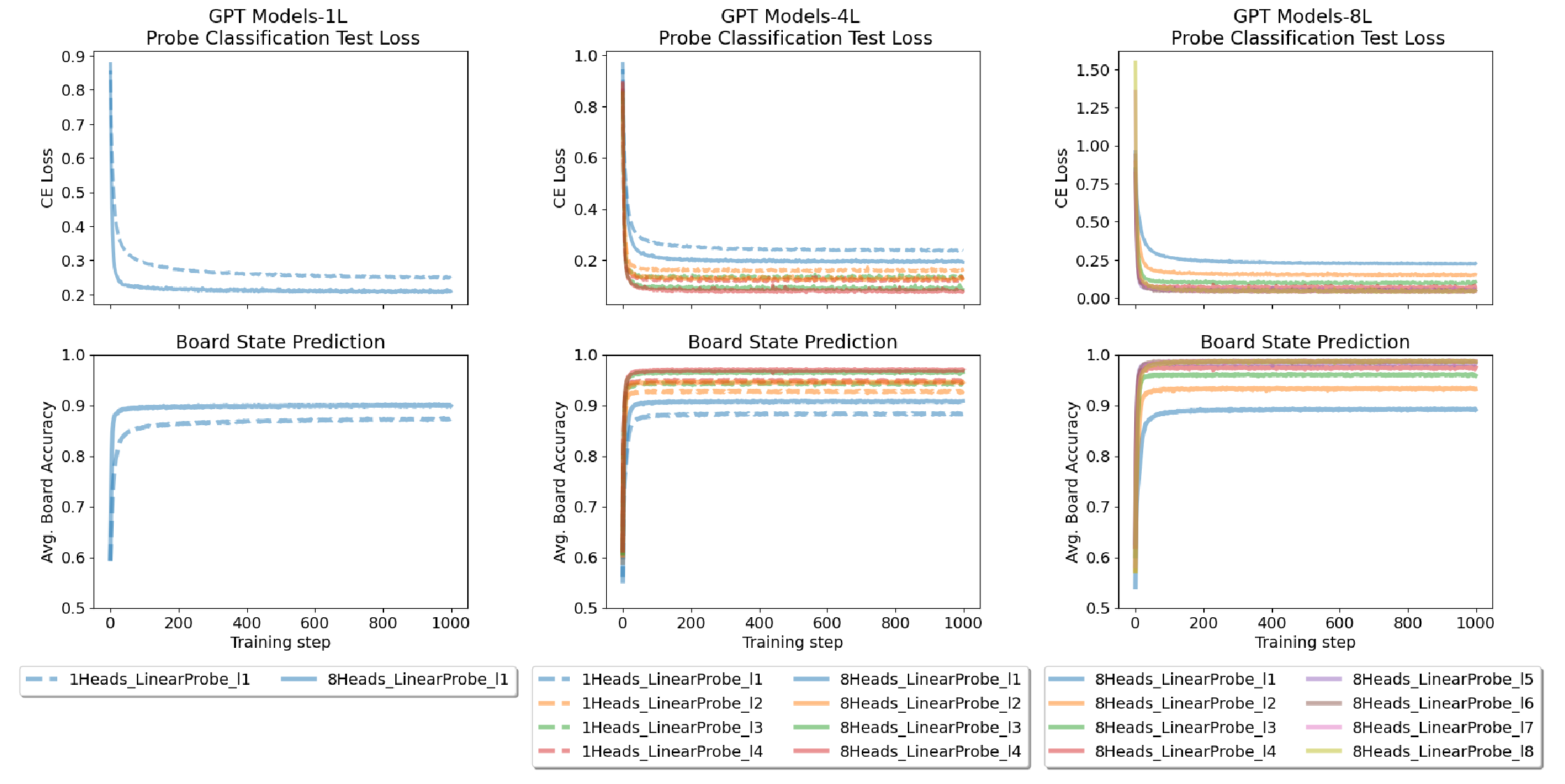}
    \caption{(First Row) Losses for various sequential models trained to play legal moves in Othello. The average percent of legal next-moves played on a validation set are [94.9$\%$, 98.6$\%$, 99.7$\%$, 99.6$\%$, 99.9$\%$] for the models in the legend. (Bottom rows)  Linear probe accuracy for different models at each layer.}
    \label{fig:trained_probes}
\end{figure}

% \begin{figure}[t!]
%     \centering
%     \includegraphics[width=1.0\columnwidth]{Figures/gpt_complex_interventions.png}
%     \caption{Examples of new interventions involving more complicated changes to the activation vectors. Starting with an ``unplayed board'' (displayed in the top left of each set), we obtain two different game sequences, the played-move sequence and the alternate-move sequence. The interventions analysis begins by passing in the alternate-move sequence to the model but then modifying the activation vectors in the model to $\textit{trick}$ the model into believing that the played-move sequence was actually passed in. The logits with and without intervention are shown (bottom right two plots in each set) along with the ``ground truth'', i.e. the logits if the ``played move'' was passed in (top right plot of the set).}
%     \label{fig:complex_intervention_method}
% \end{figure}

\begin{figure}[t!]
    \centering
    \includegraphics[width=1.0\columnwidth]{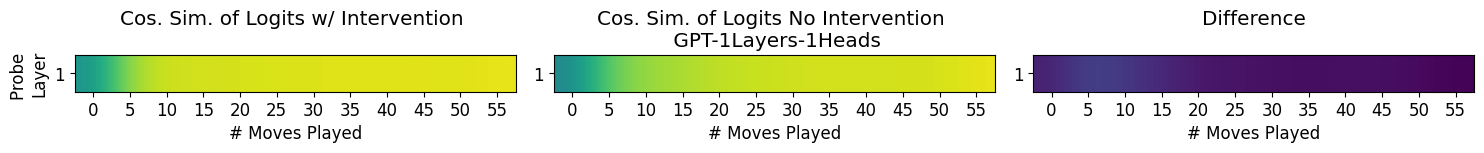}\\
    \includegraphics[width=1.0\columnwidth]{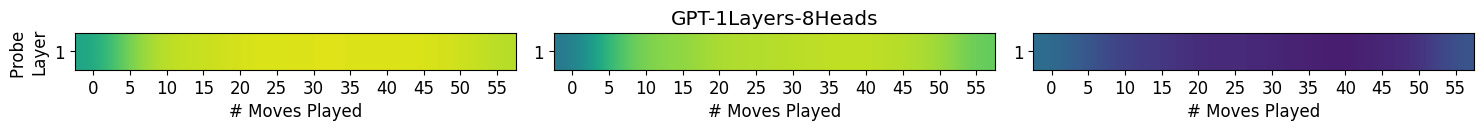}\\
    \includegraphics[width=1.0\columnwidth]{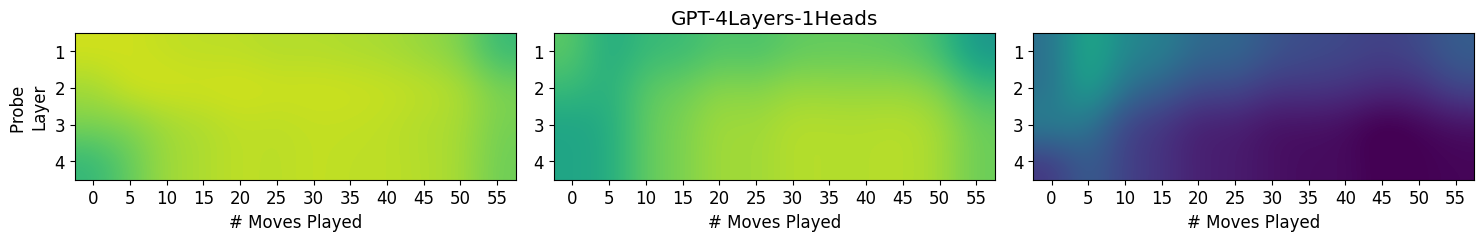}\\
    \includegraphics[width=1.0\columnwidth]{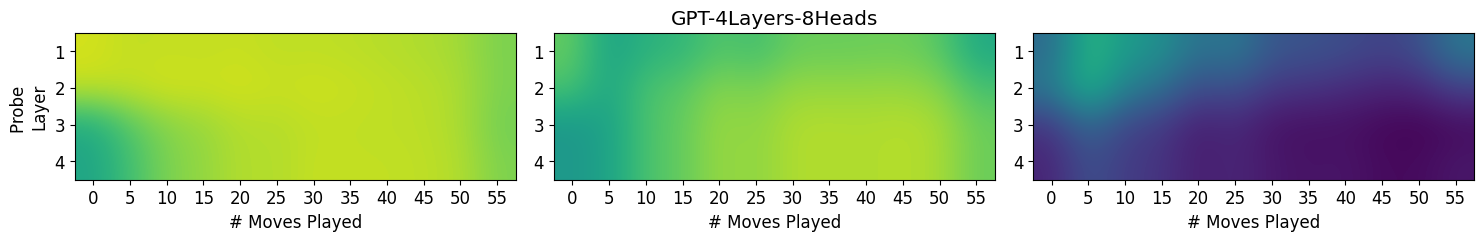}\\
    \includegraphics[width=1.0\columnwidth]{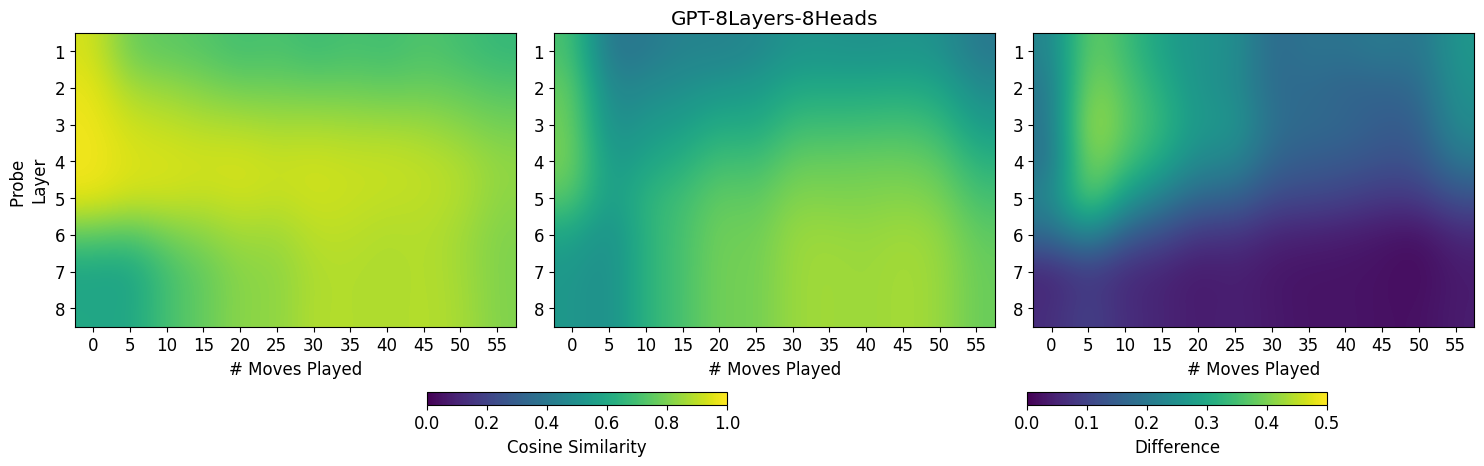}
    \caption{Cosine similarities for different Othello-GPT models with causal interventions. As in Figure \ref{fig:cos_sim} the data displayed is the cosine similarity of two logit vectors (ground truth vs. post-intervention logits or ground truth vs. pre-intervention logits), averaged over $50$ games. Interventions are conducted for different probe layers and for game sequences of different lengths. The rightmost column displays the difference between the first and second columns and reveals where/when the interventions are most impactful to the model's next-move predictions.}
    \label{fig:intervention_sweep_distance}
\end{figure}

In this section, we report our findings on linear world representations for smaller models than the 8-layer-8-head Othello-GPT.

To this end, we trained a series of shallow models of the same architecture (Figure \ref{fig:GPT_and_Probe_Schematic}a) but with a number of attention blocks (layers) between 1 and 8 and with either 1 or 8 attention heads\footnote{Our naming convention is slightly misleading. We refer to GPT models in the sense that our trained neural architectures are similar to the GPT-2 architecture. We are not utilizing ``pre-trained transformers'' since all models are randomly initialized and trained from scratch.}. The models are trained on a dataset of over 20 million simulated games (where legal moves are played but without strategy) and the training losses are shown in Figure \ref{fig:trained_probes} (top row). During training, we periodically benchmark the models performance by computing the percent of predicted next moves that are legal for a validation set (500 game sequences/29,500 move predictions). Notably, we find that even the smallest model tested, GPT 1-Layer 1-Head, learns to play legal moves approximately 95$\%$ of the time, while the larger models predict legal next-moves over 99$\%$ of the time. 

Next, we trained linear probes that map the activation vector to the new world representation for each layer and each model (see the discussion in section \ref{ssec:linear-world-representation}), with the training results shown in Figure \ref{fig:trained_probes} (bottom rows). The findings are similar to before in that the mapping is most accurate for deep layers rather than shallow (see for example the layer 1 probe vs the final layer probe). Interestingly, it appears that all Othello-GPT models tested present a similar ability to encode the world representation for a given layer; for example, the first layer probe accuracy and losses are similar for the GPT models with 1, 4, and 8 layers.

As in section \ref{ssec:causal}, we conduct causal intervention studies for these models and the results are shown in Figure \ref{fig:intervention_sweep_distance}. We see that though there is a linear representation of the board states in shallow networks (1-layer), the next-move decision-making is not using the information causally. However, the 4L1H model shows similar results as the 8L8H model, suggesting that the minimum model for causal linear world representations can be really small.

\section{Visualization of Attention Heads}
\label{appendix:8L_attn}
\begin{figure}[t!]
    \centering
    \includegraphics[width=1.0\columnwidth]{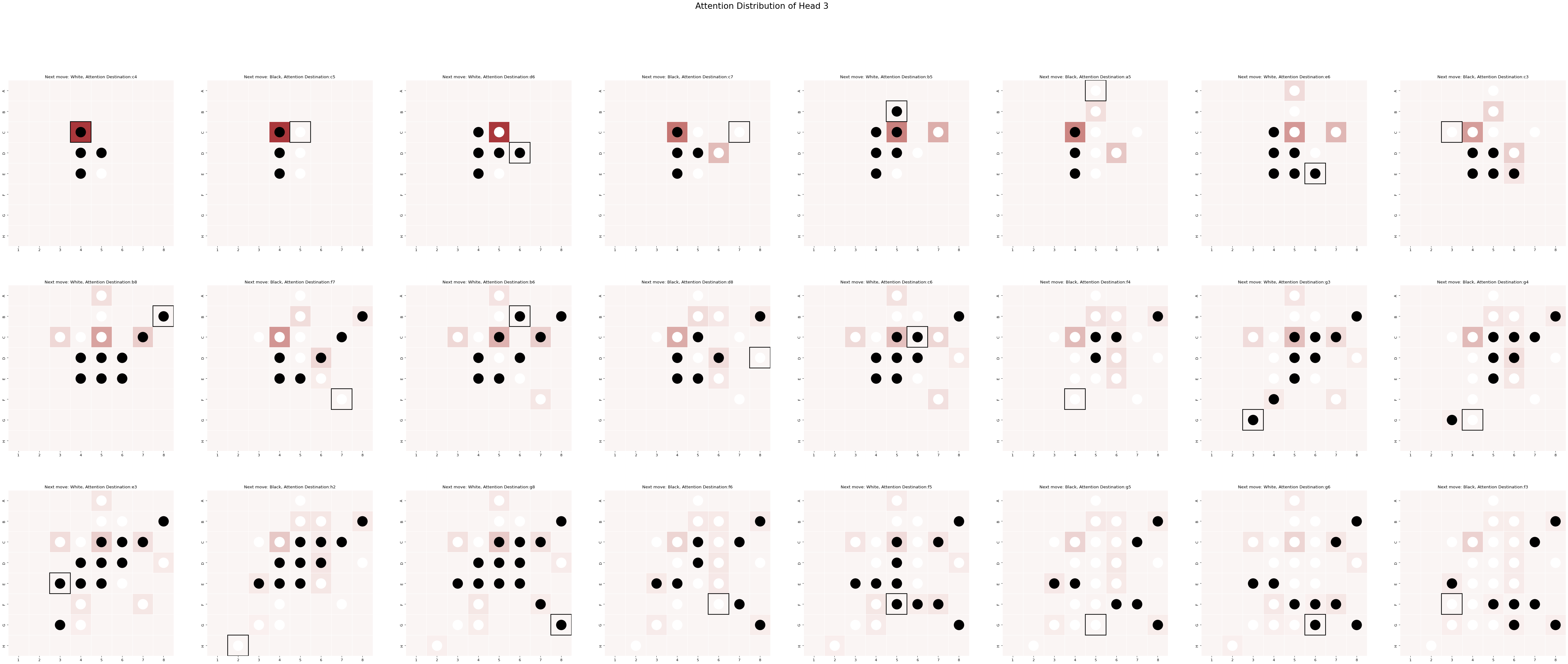}
    
    \caption{Attention distribution of head 3 on Othello board for the first 24 moves. For the same game as in Figure \ref{fig:1L8H_attn}, we show the board state and attention distribution for the last move on the board, which is labeled by the black square. Dark color corresponds to large weights and light color corresponds to small attention weights. }
    \label{fig:head3}
\end{figure}

\begin{figure}[t!]
    \centering
    \includegraphics[width=1.0\columnwidth]{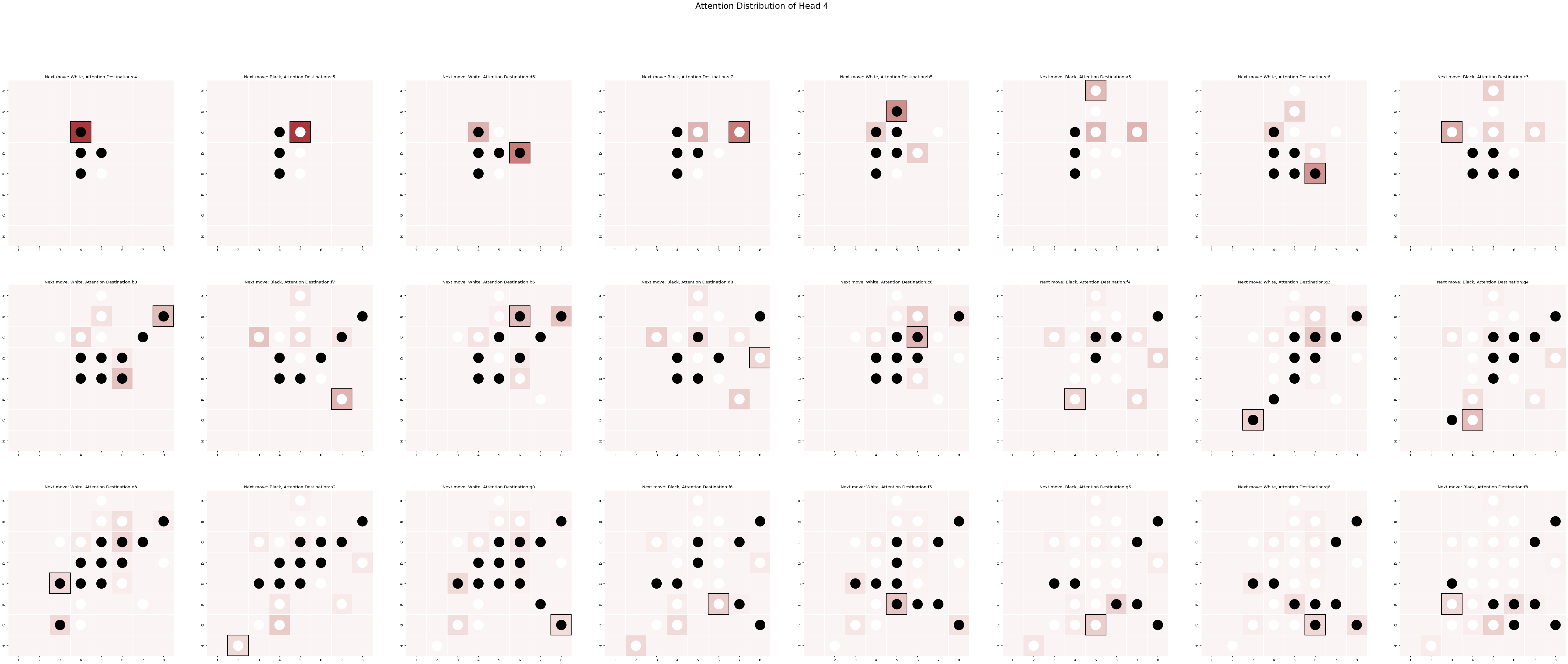}
    
    \caption{Attention distribution on Othello board for head 4. Head 4 is tracking a different set of pieces than head 3.}
    \label{fig:head4}
\end{figure}

In this section, we provide more visualizations of the attention heads in different models. As discussed in section \ref{section:att}, there exist "yours" and "mine" attention heads. We see how they keep track of the moves in Figure \ref{fig:head3} and \ref{fig:head4} respectively, as the game progresses. We see clearly that heads 3 and 4 are each tracking one-half of the game pieces, which corresponds to my and the opponent's historical moves.

\begin{figure}[t!]
    \centering
    \includegraphics[width = 1.0 \columnwidth]{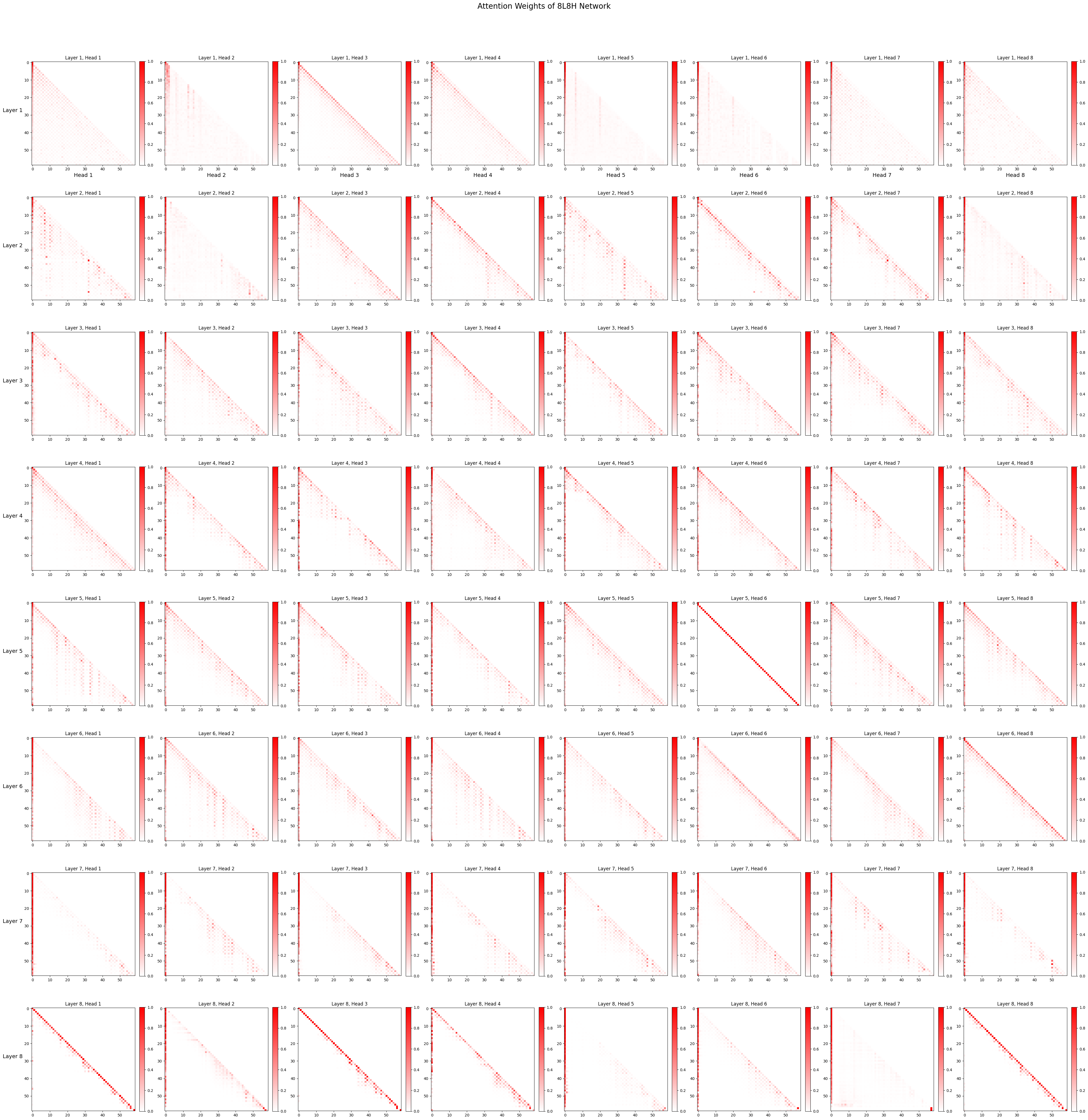}
    \caption{Attention heads for the 8L8H Othello-GPT on a sample game.}
    \label{fig:8L8H_att}
\end{figure}

\begin{figure}[t!]
\centering
\includegraphics[width=0.6\columnwidth]{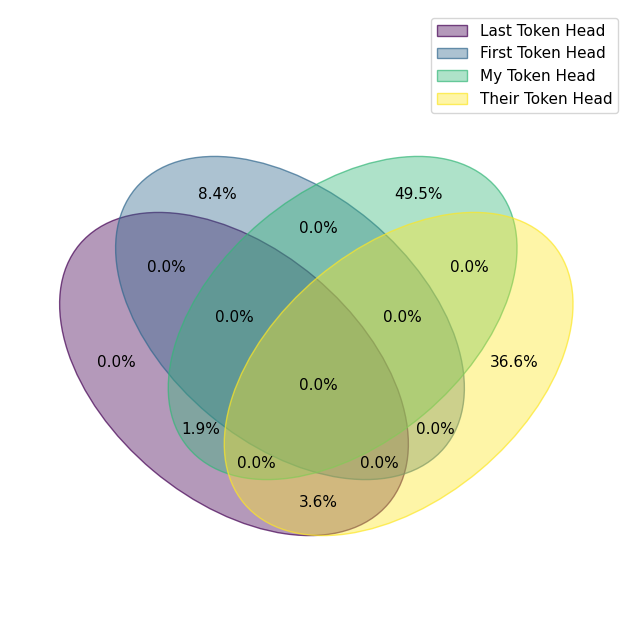} 
\caption{Venn diagram of attention head types. There are four type of circuits we find in our Othello model, last token heads (which looks at the last played move), first token heads (which look at the first played move), my token heads (which look at moves I made), and their token heads (which look at moves my opponent made).}
\label{fig:att_cat}
\end{figure}

Furthermore, we present in Figure \ref{fig:8L8H_att} all of the attention heads of the 8L8H Othello-GPT model on a sample game. They can be roughly categorized as in Figure \ref{fig:att_cat}.

The attention heads roughly fit into four categories:
\begin{table}[ht]
\centering
\caption{Distribution of Attention Head Types across Layers}
\label{table:example}
\begin{tabular}{lccccc}
\toprule
\textbf{Layer} & \textbf{"Last" Head} & \textbf{"First" Head} & \textbf{"Mine" Head} & \textbf{"Theirs" Head} & \textbf{Other} \\
\midrule
Layer 1 & 0.00\% & 0.00\% & 37.50\% & 25.00\% & 37.50\% \\
Layer 2 & 0.00\% & 0.00\% & 24.99\% & 49.98\% & 25.03\% \\
Layer 3 & 0.00\% & 0.00\% & 49.98\% & 49.99\% & 0.03\% \\
Layer 4 & 0.00\% & 0.00\% & 86.95\% & 12.50\% & 0.55\% \\
Layer 5 & 12.50\% & 0.00\% & 62.27\% & 25.00\% & 0.23\% \\
Layer 6 & 0.00\% & 0.01\% & 49.90\% & 37.50\% & 12.61\% \\
Layer 7 & 0.00\% & 28.80\% & 10.19\% & 11.54\% & 49.47\% \\
Layer 8 & 23.36\% & 25.79\% & 0.16\% & 26.49\% & 24.20\% \\
\bottomrule
\end{tabular}
\label{table: circuit_dist}
\end{table}
\begin{itemize}
    \item \textbf{First Token Heads}: These attention heads focus only on the first token of the sequence; examples include Layer 7 Head 1, Layer 6 Head 1, and Layer 8 Head 2 (\autoref{fig:8L8H_att}).
    \item \textbf{Last Token Heads}: These attention heads focus on the second-to-last (position of last move you played) or last token (position of last move your opponent played); examples include Layer 5 Head 6, Layer 8 Head 3, and Layer 8 Head 1 (\autoref{fig:8L8H_att}).
    \item \textbf{My Token Heads}: These attention heads focus on positions corresponding to moves made by the color to play (for example, if white to play, it will pay attention to moves in the sequence that were made by white). Examples include Layer 1 Head 1, Layer 2 Head 3, and Layer 3 Head 2 (\autoref{fig:8L8H_att}).
    \item \textbf{Their Token Head}: These attention heads focus on positions corresponding to moves made by the color that just played (for example, if white to play, it will pay attention to moves in the sequence that were made by black). Examples include Layer 1 Head 4, Layer 2 Head 1, and Layer 2 Head 5 (\autoref{fig:8L8H_att}).
\end{itemize}

\autoref{table: circuit_dist} shows the attention head types across layers. We see that the heads in Layers 1 through Layer 4 only consist of my token heads or their token heads. Additionally, we notice that in \autoref{fig:8L8H_att}, the "context size" of the attention heads for my token heads and their token heads seem to decrease over time despite still following the behavior we enumerated above; we see that the attention heads in earlier layers tend to pay attention to moves across the entire context, whereas attention heads in later layers only pay attention to the moves closest to them.

Another interesting finding is that the Last token heads only appear in Layers 5 and 8 and First Token Heads only in Layers 7 and Layers 8. Taking all this into account, we hypothesize that the necessary representations for the board state (my piece vs. your piece) are largely already finalized by the later layers (Layer 5 and onwards) and thus the causal interventions fail in deeper layers.
\begin{figure}[t!]
    \centering
    \includegraphics[width = 1.0\columnwidth]{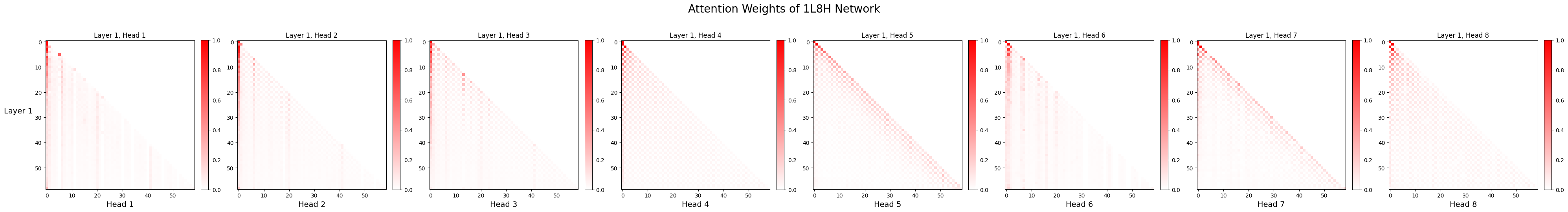}
    \caption{Attention heads for the 1L8H Othello-GPT on a sample game.}
    \label{fig:1L8H_att}
\end{figure}

Lastly, we show that "yours" and "mine" attention heads emerge even for the 1-layer transformer, shown in \autoref{fig:1L8H_att}. This is also consistent with our findings that 1L8H network has a linear representation of the board states.

%%%%%%%%%%%%%%%%%%%%%%%%%%%%%%%%%%%%%%%%%%%%%%%%%%%%%%%%%%%%
\clearpage
\bibliographystyle{ACM-Reference-Format}
\bibliography{references.bib}

\end{document}